\documentclass[journal]{IEEEtai}

\usepackage[colorlinks,urlcolor=blue,linkcolor=blue,citecolor=blue]{hyperref}

\usepackage{color,array}

\usepackage{graphicx}

\usepackage{longtable}
\usepackage[format=hang]{caption}
\usepackage{tabularray}
\usepackage{tabularx}
\usepackage{amsmath}
\usepackage{graphicx}
\usepackage{booktabs}
\usepackage{amssymb}

\usepackage{verbatim}
\usepackage[framemethod=default]{mdframed}
\usepackage{tabu}
\usepackage{multirow, makecell}
\newcommand\T{\rule{0pt}{2.9ex}}       
\newcommand\B{\rule[-1.2ex]{0pt}{0pt}} 
\usepackage{lscape}
\usepackage{tikz}
\usetikzlibrary{shapes.geometric, arrows}
\usetikzlibrary{arrows.meta, positioning}
\usepackage{pgfplots}
\pgfplotsset{compat=1.17}
\usepackage{svg}

\begin{document}

\title{Multi-Stage Retrieval for Operational Technology Cybersecurity Compliance Using Large Language Models: A Railway Casestudy} 

\author{R. Bolton, M. Sheikhfathollahi, S. Parkinson, D. Basher, and H. Parkinson
\thanks{Submitted on 17/04/2025}
\thanks{R. Bolton is with Digital Transit Limited, 3M Buckley Innovation Centre, Huddersfield, HD1 3BD, West Yorkshire, UK (e-mail: regan.bolton@digitaltransit.co.uk).}
\thanks{ M. Sheikhfathollahi is with the Department of Computer Science at University of Huddersfield, University of Huddersfield, Huddersfield, HD1 3DH, West Yorkshire, UK (e-mail: mohammadreza.sheikhfathollahi@hud.ac.uk).}
\thanks{ S. Parkinson is with the Department of Computer Science at University of Huddersfield, University of Huddersfield, Huddersfield, HD1 3DH, West Yorkshire, UK (e-mail: s.parkinson@hud.ac.uk).}
\thanks{D. Basher is with Digital Transit Limited, 3M Buckley Innovation Centre, Huddersfield, HD1 3BD, West Yorkshire, UK (e-mail: dan.basher@digitaltransit.co.uk).}
\thanks{H. Parkinson is with Digital Transit Limited, 3M Buckley Innovation Centre, Huddersfield, HD1 3BD, West Yorkshire, UK (e-mail: hjparkinson@digitaltransit.co.uk).}}


\maketitle
\begin{abstract}
Operational Technology Cybersecurity (OTCS) continues to be a dominant challenge for critical infrastructure such as railways. As these systems become increasingly vulnerable to malicious attacks due to digitalization, effective documentation and compliance processes are essential to protect these safety-critical systems. This paper proposes a novel system that leverages Large Language Models (LLMs) and multi-stage retrieval to enhance the compliance verification process against standards like IEC 62443 and the rail-specific IEC 63452.
We first evaluate a Baseline Compliance Architecture (BCA) for answering OTCS compliance queries, then develop an extended approach called Parallel Compliance Architecture (PCA) that incorporates additional context from regulatory standards. Through empirical evaluation comparing OpenAI-gpt-4o and Claude-3.5-haiku models in these architectures, we demonstrate that the PCA significantly improves both correctness and reasoning quality in compliance verification.
Our research establishes metrics for response correctness, logical reasoning, and hallucination detection, highlighting the strengths and limitations of using LLMs for compliance verification in railway cybersecurity. The results suggest that retrieval-augmented approaches can significantly improve the efficiency and accuracy of compliance assessments, particularly valuable in an industry facing a shortage of cybersecurity expertise.
\end{abstract}

\begin{IEEEImpStatement}
This research addresses a critical gap in railway cybersecurity by introducing an AI-assisted compliance verification system. As critical infrastructure becomes increasingly digitized, railways face growing cybersecurity threats that could compromise safety and operations. Manual compliance verification is time-consuming, resource-intensive, and requires specialized expertise that is increasingly scarce in the industry. Our multi-stage retrieval system using Large Language Models demonstrates significant improvements in compliance assessment quality, with our Parallel Compliance Architecture (PCA) achieving 0.1 points higher correctness scores and 0.55 points higher reasoning scores than the baseline approach as seen in Table \ref{results}. These improvements could substantially reduce verification time. 
\end{IEEEImpStatement}


\section{Introduction}

\label{sec:intro}
\IEEEPARstart{C}{ybersecurity} in Operational Technology (OT) remains a significant challenge, especially in critical infrastructure such as railways, where digital systems are used to monitor and control operations~\cite{kour2023review}. As the rail industry increasingly embraces digitalisation~\cite{jagare2019change}, the attack surface has expanded, leading to a rise in cybersecurity threats~\cite{rekik2018cyber}. Without proper management of OT cybersecurity (OTCS), safety-critical rail systems can be compromised, leading to operational failures and safety hazards~\cite{sabaliauskaite2015aligning}. Until recently, rail systems operated largely within restricted environments, isolated from wide-area network communication. However, recent technological advances, combined with an increasing demand for operational data, have led to a rapid expansion in connectivity. As a result, many safety critical systems are now exposed to networked environments for which they were not originally designed. These systems, often developed decades ago with lifespans of 30 to 40 years, did not incorporate cybersecurity considerations, exposing them to new attack vectors that were outside of the scope of their initial design and deployment. This challenge is not unique to the rail sector; however, rail has been shown to be behind other industries in cybersecurity, such as aviation, making it especially vulnerable to attack~\cite{kour2019emaintenance}. Furthermore, due to the age of some systems, the experience required to adequately assess the deployed technology may no longer be available. Therefore, there is a need to assist individuals in the absence of specialist knowledge in performing compliance exercises, which involve reviewing documentation against standards to ensure it is secure.

Effective cybersecurity documentation, development, and review processes are essential parts of the compliance process and are required to proactively protect these systems from malicious attacks. The importance of cybersecurity has resulted in dedicated international standards to ensure that the rail industry is adequately protected. Failure to adhere to such cybersecurity standards can result in significant vulnerabilities that can be exploited by an adversary. Therefore, adherence to the OT standards of IEC 62443~\cite{isa2024iec} and the anticipated rail-specific IEC 63452~\cite{bsiec63452} is crucial. These standards outline the necessary processes to achieve compliance, ensuring that organisational cybersecurity processes are established from the outset and that effective cybersecurity activities are carried out throughout the development lifecycle of any railway asset.

Compliance with these standards is a considerable challenge due to their complexity and specialised subject knowledge, making it increasingly difficult to meet regulatory requirements~\cite{tunggal2024compliance}. The process of analysing the compliance of cybersecurity documentation with regulations and standards is typically performed by someone who has extensive knowledge of the domain and familiarity with the associated standards. In general, cybersecurity assessment methods are based on
testing, examination, or interview~\cite{stewart2011cissp}. Examination can be difficult due to the inherent ambiguity in the OTCS documentation. This complexity adds to the time and cost involved in conducting a thorough compliance assessment. These issues contribute to the motivation for the presented research.

The challenges of maintaining compliance are exacerbated as any change to a train's digital systems will have an impact on the OTCS. This impact will be assessed in terms of both the new vulnerabilities and risks introduced by the new technology and the impact of the change on existing legacy systems. Therefore, it is necessary to thoroughly assess compliance to maintain a high level of security. In addition to the need to maintain compliance with the introduction of new systems, there is also the need to periodically perform an assessment to account for the evolution of cybersecurity standards. 

There is clearly a need to assist in compliance exercises; otherwise, it is likely that an organisation might fall behind with their compliance due to the knowledge and time-intensive process. Researchers have previously worked on developing intelligent compliance-checking solutions. However; the work is in different domains, such as construction~\cite{amor2021promise} and software engineering~\cite{malsane2015development}. Many solutions use forms of Artificial Intelligence, most prominently natural language processing (NLP)~\cite{zhang2016semantic}. These works provide promising results; however, they often lack the flexibility to handle a diverse domain. Large Language Models (LLMs) have recently demonstrated promise in other knowledge-intensive and time-intensive tasks. To the best of the authors' knowledge, the use of LLMs to improve automated OTCS compliance has not yet been investigated.

LLMs excel at understanding large volumes of text and can generate human-like text based on patterns learnt from their vast amount of training data\cite{hore2024llms}. A recent example demonstrates their ability in multimode industrial diagnostics~\cite{jose2024advancing}, student performance assessment~\cite{oh2024language}, and safety case generation~\cite{sivakumar2024prompting}. This capability enables them to grasp context, semantics, and nuances in human language effectively. With the introduction of increasingly powerful LLMs such as GPT-4 \cite{achiam2023gpt} and Meta's open-source LLama-3 model \cite{dubey2024llama}, these models are becoming increasingly accurate and capable of demonstrating human-like reasoning. During the past year, advances in LLMs have prompted a shift in research focus from generative capabilities to reasoning abilities \cite{zhang2024llm}.

In this paper, a system is proposed that utilises LLMs and other advanced techniques to accelerate the compliance verification process. This has resulted in the following novel contributions:

\begin{itemize}
    \item Development and evaluation of a multi-stage LLM retrieval system designed for compliance verification in the OTCS domain;
    \item A comparison of correctness and reasoning between the Baseline Compliance Architecture (BCA) and the two Parallel Compliance Architecture (PCA) retrieval system architectures; and
    \item Established metrics for correct responses, logical reasoning, and hallucination for both architectures, highlighting their strengths and weaknesses.
\end{itemize}



The remainder of this paper is structured as follows. Section~\ref{sec:related_work} describes studies related to the research presented in this paper. Section~\ref{sec:methodology} presents the methodology, explaining the retrieval system, and discussing the two architectures developed and tested in this paper. 
Section~\ref{sec:setups_results} describes the experimental setup.
Section~\ref{sec:results} 
presents the results.
Section~\ref{sec:res_eval} provides an analysis of the observations made from the dataset. Furthermore, this section includes a discussion and the limitations of the study, and Section~\ref{sec:conclusion} concludes the paper with a conjecture and discusses avenues for future work.

\section{Related work}
\label{sec:related_work}
LLMs are being used in multiple sectors for tasks that involve the analysis of textual artefacts in different ways. In many cases, researchers employ an ensemble of LLM techniques in many domains to enhance performance and accuracy~\cite{perak2024incorporating, bianchini2024enhancing}. However, these techniques are not yet fully exploited in the OTCS domain.

One new technique in such research is retrieval augmented generation (RAG)~\cite{lewis2020retrieval} which combines the strengths of traditional information retrieval systems (such as databases) with the capabilities of generative large language models. By combining this additional knowledge with its own language skills, AI can write text that is more accurate, up-to-date, and relevant to the individual's specific needs~\cite{google2024rag}. This is powered by a retrieval engine that works by efficiently retrieving informational nodes from an external database, using a retriever, and then incorporating this context into the context window of the LLM.

Research has already addressed compliance in the architecture, engineering and construction industry (AEC)~\cite{dimyadi2013automated}. One such approach focused on evaluating different prompt engineering techniques \cite{liu2023gpt}. Their compliance goal was much simpler than that of this project. More specifically, the authors used pairs of fire safety regulations and building design specifications. Their conclusions determined that the design of prompt engineering largely determined the performance of the LLM. In this research, detailed prompts are constructed to instruct the LLM. Despite their small data set and less complex domain, LLMs showed promise in classifying compliance and non-compliance. The disadvantage of using long context is that it may not perform well with larger specifications. Regulations are likely to target key areas of the design specifications and, as a result, the context may become diluted. This dilution can make it challenging for the system to focus on the most relevant information, potentially reducing the accuracy and effectiveness of compliance verification.

An example, in the medical field, involves an investigation of how compliance can be assessed against reporting guidelines in clinical trials \cite{wrightson2023gpt}. They manually extracted text-question pairs as their method. The findings showed that the LLM demonstrated an acceptable classification accuracy of greater than 95\% in its compliance evaluation. They attribute this success to the fine-tuning of their models as increased performance is observed when this occurs. Although fine-tuning the model is not performed, a lightweight solution can be applied using a retriever to add further context and enhance the system's ability to provide correct answers. One conclusion from their paper is that it is likely that an analysis of all the papers (standards) will be required to further improve performance. The presented solution to this issue involves using an additional retriever on all of the user documentation. This enables the control of the size, amount, and overlap of nodes, which means that only specific parts of the documentation are retrieved.

Another study focuses on how assistive technology can be analysed for compliance with product specification standards \cite{arora2024towards}. Their approach is unique in the way that they use a retriever to trace product specifications to the relevant standards, then in a second stage analyse compliance using an LLM with manually inserted rules for compliance. The benefit of this approach is that it is more specific than other research, enabling the system to focus on the most relevant information. This multi-stage approach to identifying compliance is similar to the presented architecture, except that the system uses RAG to generate a contextualised response, automatically inferring the compliance rules.

Due to the different methods such as prompt engineering, retrieval augmented generation, and fine-tuning, it is likely that the research will eventually involve an ensemble of these methods. This is supported by the fact that complex domains, such as OTCS compliance, will have a wide variety of different types of documentation and complex rules that must be followed. At this preliminary stage, fine-tuning will not be performed as it has a large overhead and can be performed at the end of experimentation. However, other methods will be used. One downside to every technique is that, in one way or another, the solution is not easily generalisable or involves a manual separation or analysis. This research is uniquely applied to rail OTCS; however, the automation of the proposed system means that it can be deployed to any domain, with minor modification.

\section{Methodology} 
\label{sec:methodology}
In order to develop a system to use LLMs to assist with compliance verification, the following multi-stage methodology is presented. Several parameters are used in various presented prompt templates, these are explained in Table \ref{table_13}. In this section, the concepts of RAG retrieval, OTCS standards, BCA, and PCA are defined and discussed. 

\begin{table}[t]
\caption{A table of parameters used in prompt templates and their meaning}
\label{table_13}
\centering
    {\tabulinesep=0mm
        \begin{tabu}{l|p{6.5cm}}
            \hline\hline
            Parameter &
            Explanation \T\B\\
            \hline\hline
            \T \textit{query\_str} & The input question for the \it {input component.} \T\B\\
            \textit{user\_docs\_str} & A string of retrieved-context nodes from the \it{document retriever.}\\
             \textit{context\_str} & A string of retrieved-context nodes from the \it {context retriever.}\B\\
                      
            \hline\hline
        \end{tabu}
    }
\end{table}

\subsection{RAG retrieval}
\label{sec:methodology:RAG}

To answer questions about documentation, there are several options. One approach would be to use long context; however, in this paper's use case, which involves a massive volume of data, this would drastically increase input token usage and reduce precision. LLMs are known to `forget' information when processing large context windows, often prioritising details at the beginning and end of token sequences while under-representing or overlooking information in the middle~\cite{li2024long}. A more efficient solution is to add context to the context window during inference. Instead of manually adding document chunks, a retriever can be leveraged to automatically retrieve relevant document chunks at inference time. RAG retrieval has already proven to be very powerful in improving generation quality by adding additional context \cite{salemi2024evaluating}. However, it is proposed that RAG can also be used to retrieve relevant information from lengthy OTCS documentation and answer targetted questions about documents.

In the presented RAG model, the retriever selects the most relevant \(K\) documents, and the generator uses these documents to create a probability distribution to generate the next output token. Specifically, the input query is denoted as \(x\), and the retrieved document as \(y\), which is used as an additional context to generate the target as \(z\), the RAG retriever model is composed of two core components. 
\begin{enumerate}
    \item Retriever \(p_\eta(y|x)\), This component, parametrised by \(\eta\), retrieves a set of relevant documents based on the input query \(x\). This provides a distribution over textual resources and returns the top \(K\) documents that are most likely to contain useful information to answer the query.
    \item  Generator \(p_\theta(z_i | x, y, z_{1:i-1})\), The generator, parametrised by \(\theta\), predicts the next token \(z_i\) in the target sequence \(z\). Based on this components, the probability of generating the target \(z\) given the query input \(x\) and the retrieved document \(y\) can be described as:
    
\begin{equation}
p_{\text{RAG}}(z | x) = \prod_{i}^{N} \sum_{y \in \text{top-K}(p(\cdot | x))} p_\eta(y | x) \, p_\theta(z_i | x, y_i, z_{1:i-1})
\end{equation}

\end{enumerate}

The retrieval component \(p_\eta(y|x)\) in the presented system uses a hybrid approach, which combines multiple types of retrieval to improve the retrieval process and a vector store containing embeddings from embed-english-v3.0 embedding model is used~\cite{cohere2024}. This retriever uses the hybrid query mode, enabling it to combine both vector-based similarity (cosine similarity) and keyword-based similarity (BM25) for more precise retrieval results.
The retriever first retrieves the top-K = 10 documents using a hybrid retrieval approach, These top-10 documents are then ranked using a Cohere re-ranking model, and the top-K = 2 documents are selected for further processing. This two-step process ensures a more meaningful alignment with the input query
\(x\), leveraging both hybrid retrieval and advanced re-ranking to provide highly relevant results for the generator.

\begin{equation}
\begin{split}
S(x, y) &= \alpha \cdot \text{Cosine Similarity}(x, y) \\
&+ (1 - \alpha) \cdot \text{BM25 Similarity}(x, y)
\end{split}
\end{equation}

where \(x\) is the input query and \(y\) is the retrieved document. 
\(\alpha\) ranging from 0 to 1, controls the balance between cosine similarity and BM25. 
The retriever is parametrised with an alpha value of \(\alpha=0.75\), which controls the balance between the vector search and the textual search. This value \(\alpha=0.75\) has been generally tested and is considered the best practice in optimising retrieval performance. 

 The retrieved documents \(y\) are then used as context by the generator to produce the target sequence \(z\).



To the best of the authors' knowledge, this is a novel approach to combine RAG and OTCS for this use case. Using these methods, the aim is to improve the efficiency of cybersecurity assessments by allowing OTCS assessors to easily be able to query extensive complex OTCS documentation. These questions will be largly based on the latest rail OTCS standard IEC6345~\cite{bsiec63452}. Considering that relevant information can be spread across multiple documents, it is expected that this system will significantly enhance the quality of the results by effectively retrieving precise relevant information based on the input query.

\subsection{Baseline compliance architecture (BCA)}
\label{sec:methodology:Baseline}

The presented methodology leverages the ability of a large language model (LLM) to interpret OTCS compliance within the rail domain. When using an LLM to address OTCS-compliance with long context document processing, the responses tend to lack precision and detail. However, by targetting queries with RAG, it is expected to significantly improve the quality of the analysis. Henceforth, the RAG retriever used for this purpose shall be referred to as the \textit{document retriever}. Specifically, the retriever acts on a vector index containing the chunked internal OTCS case study.

\begin{figure}[h]
  \centering
  \begin{tikzpicture}[
    scale=0.8, 
    transform shape,
    node distance=1.5cm, 
    every node/.style={fill=red!60, font=\sffamily}, 
    align=center
  ]
    \node (input) [draw, rounded corners] {Input\\Component};
    \node (prompt_template) [draw, rounded corners, below right=0.5cm of input] {Prompt\\Template};
    \node (ud_retriever) [draw, rounded corners, below = of input] {Document\\ Retriever};
    \node (llm) [draw, rounded corners, right=of prompt_template] {LLM};
    \node (output) [draw, rounded corners, right=of llm] {Output};
    \draw[->] (input) -- (ud_retriever);
    \draw[->] (ud_retriever) -- (prompt_template);
    \draw[->] (input) -- (prompt_template);
    \draw[->] (prompt_template) -- (llm);
    \draw[->] (llm) -- (output);
  \end{tikzpicture}
  \caption{Flowchart of the basic RAG system architecture.}
  \label{fig_1}
\end{figure}
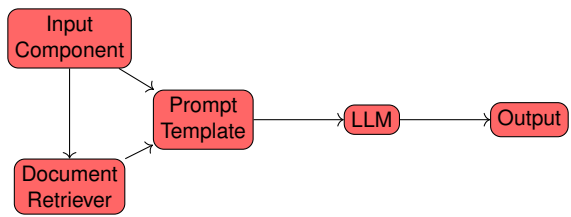



To use and test this system, the architecture as depicted in Figure~\ref{fig_1} is created. The prompt template in Figure \ref{fig:user_bca} merges the \textit{query\_str} with the retrieved nodes (\textit{document\_retriever}) and adds a system prompt in Figure \ref{fig:system_bca} for the LLM.\footnotemark[1]
\footnotetext[1]{Note: Throughout the paper, the new lines in the code block have been modified for readability.}

\begin{figure*}
\begin{mdframed}
\begin{verbatim}
You are an AI assistant specialized in reviewing documentation.
Your primary task is to perform an expert analysis on the **User Documentation**.
**Do NOT** use any prior knowledge.
Your analysis should be detailed and based directly on evidence from the 
**User Documentation**.
\end{verbatim}
\end{mdframed}
\label{fig:system_bca}
\caption{System prompt for the BCA}
\end{figure*}

A detailed system prompt is used to increase the likelihood that the LLM will follow instructions accurately, avoid potential confusion, and generate contextually relevant high-quality responses. This structured approach improves the likelihood that the model will remain focused on the task and provide accurate answers. 

\begin{figure*}
\begin{mdframed}
\begin{verbatim}

You will be provided with some documentation.

===================== **User Documentation** =====================
{user_docs_str}
==================================================================

Based **solely** on the **User Documentation**, please answer the
following **Question**.

**Question:** {query_str}
**Important Guidelines:**
- **Do NOT** use any prior knowledge or external information.
Your response **must** be in the following format:
- First provide step-by-step reasoning on how to answer the **Question**
- Then provide a summary of how you reached your answer.
\end{verbatim}
\end{mdframed}
\label{fig:user_bca}
\caption{User prompt for the BCA}
\end{figure*}

\subsection{Parallel compliance architecture (PCA)}
\label{sec:methodology:parallel}

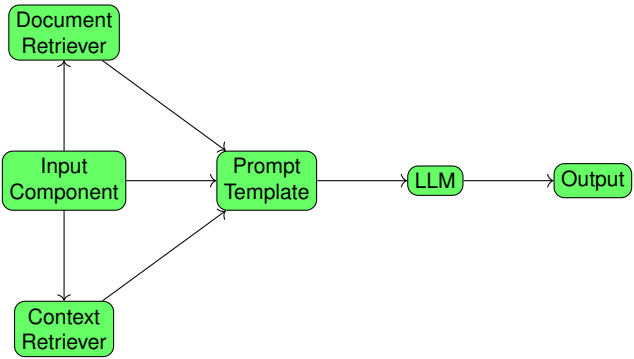
\begin{figure}[h]
  \centering
  \begin{tikzpicture}[
    scale=0.8, 
    transform shape,
    node distance=1.5cm, 
    every node/.style={fill=green!60, font=\sffamily}, 
    align=center
  ]
    \node (input) [draw, rounded corners] {Input\\Component};
    \node (prompt_template) [draw, rounded corners, right=of input] {Prompt\\Template};
    \node (dk_retriever) [draw, rounded corners, below=of input]{Context\\ Retriever};
    \node (ud_retriever) [draw, rounded corners, above=of input] {Document\\ Retriever};
    \node (llm) [draw, rounded corners, right=of prompt_template] {LLM};
    \node (output) [draw, rounded corners, right=of llm] {Output};
    \draw[->] (input) -- (dk_retriever);
    \draw[->] (input) -- (ud_retriever);
    \draw[->] (dk_retriever) -- (prompt_template);
    \draw[->] (ud_retriever) -- (prompt_template);
    \draw[->] (input) -- (prompt_template);
    \draw[->] (prompt_template) -- (llm);
    \draw[->] (llm) -- (output);
  \end{tikzpicture}
  \caption{Flowchart of the parallel system architecture.}
  \label{fig_2}
\end{figure}


To improve the models understanding of the OTCS domain, one solution could be to perform fine tuning of a model; however, this is time consuming, expensive, and does not integrate easily with retrieval augmented systems. Instead, we define another retriever that uses exactly the same methods as the document retriever, except on a different vector index. This process is named the {\it context retriever}. The context retriever returns OTCS standards and regulation information. This allows the system to improve the reasoning process and understanding of queries, using additional regulatory context. In this work, the full IEC 62443 (1-1 to 4-2) and IEC 63452 are used as material.



The parallel architecture is depicted in Figure \ref{fig_2}. The main difference is the addition of the context retriever and the prompt engineering techniques used to make the LLM understand the additional data.

\begin{figure*}
\begin{mdframed}
\begin{verbatim}
You are an AI assistant specialized in reviewing documentation based 
on the provided User Documentation.

Your primary task is to perform an expert analysis on the 
**User Documentation** using the provided **Contextual Information** 
to enhance your analysis where appropriate and necessary.
**Do NOT** use any prior knowledge or perform your analysis directly 
on the **Contextual Information**; it is provided **ONLY** to help you 
understand the **Question** and enhance your reasoning capabilities.

Your analysis should be detailed and based directly on evidence fom the 
**User Documentation**.

\end{verbatim}
\end{mdframed}
\label{fig:system_pca}
\caption{System prompt for the PCA}
\end{figure*}

In summary, the system prompt in Figure \ref{fig:system_pca} instructs the LLM to respond to a question, referencing the user documentation. Furthermore, the \textit{query\_str} is rewritten to include the \textit{user\_docs\_str} and \textit{context\_str} as seen in Figure \ref{fig:user_pca}.

\begin{figure*}
\begin{mdframed}
\begin{verbatim}
You will be provided with some documentation and supporting context:

===================== **User Documentation** =====================
{user_docs_str}
==================================================================

------------------- **Contextual Information** -------------------
{context_str}
------------------------------------------------------------------
Based **solely** on the **User Documentation** and by enhancing your 
analysis utilising the **Contextual Information**
please answer the following question.

**Question:** {query_str}

**Important Guidelines:**
- **Do NOT** use any prior knowledge or external information.
- **Do NOT** perform an analysis of the **Contextual Information** 
in your answer.
Your response **must** be in the following format:
- First Provide step-by-step reasoning on how to answer the **Question**, 
potentially making use of the **Contextual Information** to refine your 
steps.
- Then provide a summary of how you reached your answer.

\end{verbatim}
\end{mdframed}
\label{fig:user_pca}
\caption{User prompt for the PCA}
\end{figure*}



\section{Experiment}
\label{sec:setups_results}

This section describes the testing setup used in this research and justifies the used evaluation methodology.

\subsection{Implementation details}

Throughout the tests, GPT-4o \cite{wiggers2024openai} and Claude-3.5-Haiku \cite{rahman2024automatic} are used to generate responses,\cite{dubey2024llama} with the Llama-3.1-405B-Instruct model used as an evaluator. All models have a max\_tokens setting of 2048. All models are accessed using an API, provided by Openrouter~\cite{openrouter_ai}. Similarly, the Cohere embed-english-v3.0 model is used for the embeddings~\cite{cohere2024}.

LlamaIndex \cite{Liu_LlamaIndex_2022} for Python \cite{van1995python} is used to connect custom data sources to the LLMs. The nodes are created with a chunk size of 1024 tokens and a chunk overlap of 20 tokens in both vector indexes described in Section \ref{sec:methodology}. The retrieval mechanism described in Section~\ref{sec:methodology:RAG} is used for both retrievers, context, and documentation. LlamaParse, a genAI-native parsing platform, is used to preprocess OTCS documentation, including user documents and standards. Built for LLM use cases, LlamaParse ensures high-quality data through advanced features like table extraction and multi-file-type support, optimising the retrieval process \cite{Liu2024}. This is essential for complex OTCS documentation which can be very large and include tables for things like requirements.

\subsection{Dataset description}

To test the approach, queries from a dataset of 44 questions covering various aspects of OTCS compliance are used, primarily using the model GPT-4o to analyse and respond to the questions (unless specified). The questions were designed with varying levels of difficulty and diversity to challenge the system. Primarily, the questions were taken from the new IEC63452 standard \cite{bsiec63452}. The expected answers included both compliant, non-compliant and partially compliant scenarios. 

\subsection{Evaluation methodlogy}

The evaluation method throughout the experimentation consisted of a mixture of human- and LLM-based evaluations. Given the complexity of the OTCS domain, it is necessary to involve a human expert in OTCS. Recent work in legal analysis~\cite{cheong2024not} and security operation centre analysis~\cite{oniagbi2024evaluation} both use expert evaluation. Building on this, in the presented approach, the expert must decide whether the answer is satisfactory by responding with 'Correct,' 'Not Correct,' or 'Partially Correct.' The expert should also provide a rationale for their decision. Additionally, the clarity of the reasoning in the answers is assessed by responding with 'Strongest,' 'Weakest,' or 'Moderate', evaluated comparatively between the three tested architectures, to ensure a comprehensive evaluation of both accuracy and coherence between them. The human evaluation methodology used aligns with the research's aim; specifically, the aim is to increase the confidence in compliance verification by improving reasoning and correctness as much as possible.

For LLM-based evaluation, a technique called LLM-as-a-judge~\cite{zheng2023judging} is used, which prompts an LLM to assess another LLM's response based on predefined criteria. In this work, the focus is on evaluating correctness and reasoning ability while also analysing the impact of retrieved context on response quality. The evaluation was carried out using the Arize-Phoenix tool~\cite{arize_ai_phoenix}.  Recent research has demonstrated that LLMs tend to exhibit bias such as self-recognition and self-preference, when evaluating their own responses~\cite{panickssery2024llm}.Furthermore, evidence suggests that larger models outperform smaller ones in evaluation tasks~\cite{thakur2024judging}, prompting us to use Llama-3.1-405B-Instruct as the evaluation model~\cite{dubey2024llama}.

In the presented evaluation process, the focus is on hallucination detection and identifying instances where the model generated factually incorrect or hallucinated information. This is motivated by the research aim of increasing the confidence of LLM systems. Other LLM-based evaluation was considered; however, it required expert subject knowledge or was too subjective to be reliably accurate. 

\section{Results}
\label{sec:results}

\subsection{BCA results}

\begin{figure}
    \centering
    \includegraphics[width=0.60\linewidth]{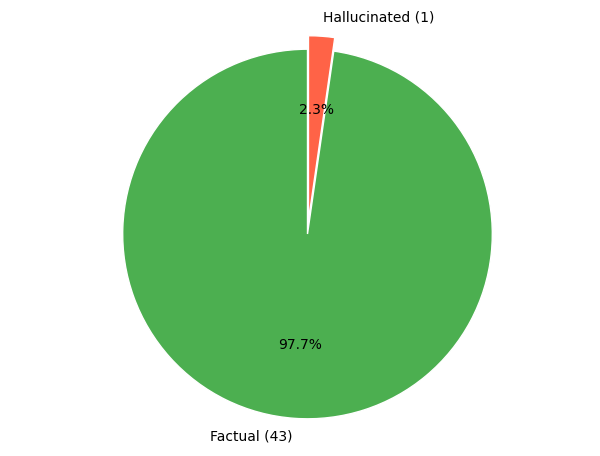}
    \caption{Evaluation of BCA Hallucination by LLM: Factual vs. Hallucinated Answers, Using the Llama-3.1-405B-Instruct model.}
    \label{BCA_Hall}
\end{figure}



After generating responses from the dataset, the hallucination evaluation for the BCA is presented in Figure~\ref{BCA_Hall}. Of the 44 questions tested, only one (2.3\%) exhibited a hallucinated response. This is a positive outcome, emphasising the effectiveness of the retrieval system in providing highly relevant and accurate information to guide the model's responses, thereby minimising the likelihood of errors commonly associated with LLMs. 

\begin{figure}
\centering
\begin{tikzpicture}[scale=0.8] 
\begin{axis}[
	x tick label style={
		/pgf/number format/1000 sep=},
	ylabel=Number of questions (44),
	enlargelimits=0.05,
	ybar interval=0.7,
    grid=major,
        symbolic x coords={Correctness, Reasoning, end},
        legend style={at={(0.5,-0.15)},anchor=north,legend columns=-1},
	legend entries={Correct/Strongest, Partially correct/Moderate, Not correct/Weakest}
]
\addplot 
	coordinates {(Correctness,19) (Reasoning,2) 
		 (end,30)};

\addplot 
	coordinates {(Correctness,17)(Reasoning,14)  
		(end,15)};

\addplot 
	coordinates {(Correctness,6) (Reasoning,26)  
		(end,4)};
\end{axis}
\end{tikzpicture}
\caption{Results of the human evaluation for BCA: correctness (left) and reasoning (right)}
\label{Human_Evaluation_BCA}
\end{figure}
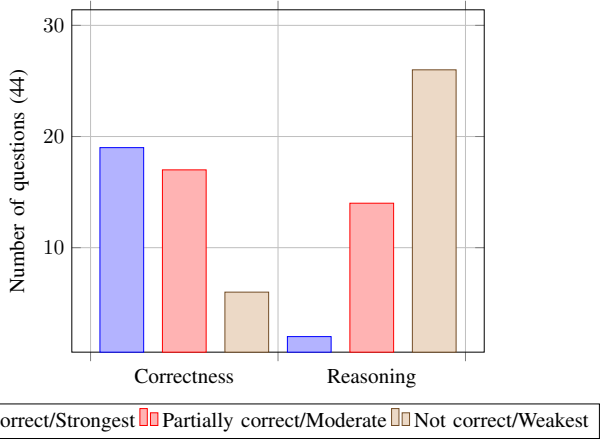

In addition, a human evaluation was conducted to further assess the quality of responses. The results of this evaluation, shown in Figure~\ref{Human_Evaluation_BCA}, provide insight into how cybersecurity experts classified the correctness of the model. This evaluation complements the automatic evaluation by offering a detailed, expert-driven perspective on the model's performance.

As shown in Figure~\ref{Human_Evaluation_BCA}-left, out of the 44 questions in the dataset, 19 were classified as correct, 17 as partially correct, and 6 as incorrect. This distribution suggests that the system can accurately answer just under half of the questions through user documentation retrieval alone, with a significant portion requiring further refinement and a smaller fraction being entirely incorrect. The ``partially correct'' responses often contained elements of accuracy, but fell short in terms of completeness or precision in addressing the specific nuances of the questions. For example, some answers provided general information relevant to OTCS, but did not capture the exact technical details or contextual specifics demanded by the question. Conversely, the ``incorrect'' responses were either factually inaccurate or entirely unrelated to the query, highlighting areas where the model's comprehension or retrieval mechanisms require significant improvement.  

Figure~\ref{Human_Evaluation_BCA}-right, shows the reasoning evaluation, where, of the same 44 questions, only 2 exhibited the strongest reasoning, 14 were moderate and 26 were weakest, highlighting a challenge in constructing well-reasoned, logically coherent responses despite retrieving relevant information. Although the system excels at avoiding hallucinations and providing relevant OTCS data, it requires additional context to improve accuracy, completeness, and reasoning quality.

\subsection{PCA experiment}
\label{subsec:PCA_Experiment}

\begin{figure}
    \centering
    \includegraphics[width=0.85\linewidth]{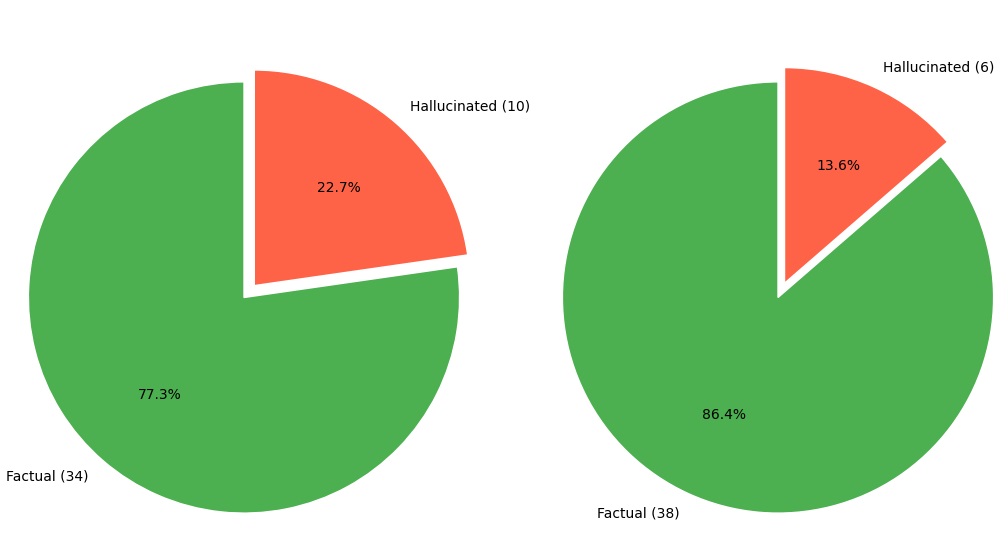}
    \caption{Evaluation of PCA Hallucination by LLM: Factual vs. Hallucinated Answers, Using the Llama-3.1-405B-Instruct Model for evaluation and GPT-4o as the response model (left) and Claude-3.5-Haiku as the response model (right).}
    \label{PCA1_PCA2}
\end{figure}

Although the BCA performs well in avoiding hallucinations and providing relevant information, it still requires more context on OTCS to improve correctness and reasoning. Based on this,  the performance of the PCA is evaluated using the same 44 
questions as the BCA. To determine whether the choice of LLM model significantly affects performance, a control experiments was conducted using Claude 3.5 Haiku as an alternative provider to GPT-4o.

The results of the hallucination evaluation are illustrated in Figure \ref{PCA1_PCA2}, which provides a visual representation of the performance of the model in terms of hallucination. The figure demonstrates that the hallucination rate remains relatively consistent across both models, suggesting that the effectiveness of the retrieval system plays a crucial role in maintaining response quality. This consistency can be attributed to the robust performance of the retrieval system, which ensures that high-quality, relevant information is retrieved for processing by the models, as the quality of retrieved data plays a critical role in shaping the output. The increase in ``hallucinated'' answers from the BCA can be explained by the additional context that is used in the PCA and the necessity of using this context directly in the answer. It is important to note that the context often wont appear directly in the response and as a result the entry is marked as ``hallucinated'' when in fact it has aided the understanding of the query for the LLM without direct reference in the response. That being said it is still useful to compare the hallucination between the two PCA despite this metric limitation. 

\begin{figure}
\centering
\begin{tikzpicture}[scale=0.8] 
\begin{axis}[
	x tick label style={
		/pgf/number format/1000 sep=},
	ylabel=Number of questions (44),
	enlargelimits=0.05,
	ybar interval=0.7,
    grid=major,
        symbolic x coords={BCA, PCA1, PCA2, end},
        legend style={at={(0.5,-0.15)},anchor=north,legend columns=-1},
	legend entries={Correct, Partially correct, Not correct}
]
\addplot 
	coordinates {(BCA,19) (PCA1,27) (PCA2,27)
		 (end,25)};

\addplot 
	coordinates {(BCA,17)(PCA1,9) (PCA2,10) 
		(end,15)};

\addplot 
	coordinates {(BCA,6) (PCA1,6) (PCA2,5) 
		(end,4)};
\end{axis}
\end{tikzpicture}
\caption{Results of the human evaluation on correctness for BCA (left) using GPT-4o, PCA1 (middle) using GPT-4o, and PCA2 (right) using Claude-3.5-Haiku.}
\label{Human_Evaluation_Correctness_BCA_PCA1_PCA2}
\end{figure}
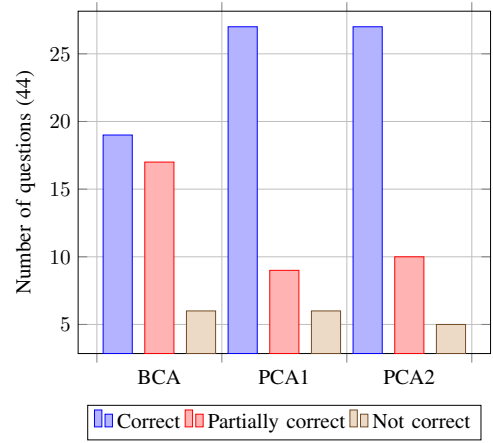

Human evaluation was also performed to gain deeper insight into the quality of the models' responses. The findings of this evaluation, carried out by a cybersecurity expert, are shown in Figure \ref{Human_Evaluation_Correctness_BCA_PCA1_PCA2}. The responses of PCA1 and PCA2 are nearly identical, confirming the effectiveness of the retrieval mechanism. Both architectures correctly answered 27 of the 44 questions, demonstrating similar performance. PCA1 provided partially correct answers for 9 questions, while PCA2 did so for 10 questions. Furthermore, PCA1 and PCA2 incorrectly answered 6 and 5 questions, respectively.

Compared to BCA, the PCA demonstrated a significant increase in the number of completely correct responses and a decrease in the number of partially correct responses. This indicates a shift toward more accurate and definitive answers.
A key reason for this improvement is that the PCA uses both retrievers, allowing the system to synthesise responses more effectively. Unlike the BCA, which relies primarily on direct retrieval from user documents, the PCA incorporates context to reach a deeper understanding, resulting in more coherent and well-reasoned responses. Interestingly, we can observe that the number of incorrect responses are relatively similar between all the models; this demonstrates that when document retrieval fails, no amount of additional context can help the model give a more correct response. This shows that document retrieval is likely a bottleneck of the system.

\begin{figure}
\centering
\begin{tikzpicture}[scale=0.8] 
\begin{axis}[
	x tick label style={
		/pgf/number format/1000 sep=},
	ylabel=Number of questions (44),
	enlargelimits=0.05,
	ybar interval=0.7,
    grid=major,
        symbolic x coords={BCA, PCA1, PCA2, end},
        legend style={at={(0.5,-0.15)},anchor=north,legend columns=-1},
	legend entries={Strongest, Moderate, Weakest}
]
\addplot 
	coordinates {(BCA,2) (PCA1,26) (PCA2,14)
		 (end,25)};
\addplot 
	coordinates {(BCA,14)(PCA1,12) (PCA2,17) 
		(end,15)};
\addplot 
	coordinates {(BCA,26) (PCA1,4) (PCA2,11) 
		(end,4)};
\end{axis}
\end{tikzpicture}
\caption{Results of the human evaluation on reasoning for BCA (left) using GPT-4o, PCA1 (middle) using GPT-4o, and PCA2 (right) using Claude-3.5-Haiku.}
\label{fig:Human_Evaluation_Reasoning_BCA_PCA1_PCA2}
\end{figure}
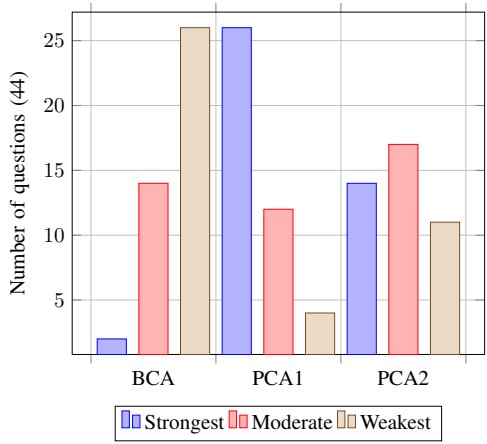

As shown in Figure~\ref{fig:Human_Evaluation_Reasoning_BCA_PCA1_PCA2}, the BCA performs significantly worse than PCA1 and PCA2 in terms of reasoning. The reasoning in PCA1, which uses GPT-4o, demonstrates the strongest performance. Among the 44 questions, PCA1 had the strongest reasoning in 26 cases, while PCA2 did so in 14. PCA1 provided moderate reasoning for 12 questions, compared to 17 for PCA2. Furthermore, PCA1 exhibited the weakest reasoning in 4 instances, whereas PCA2 did so in 11. These results demonstrate that GPT-4o is better fine-tuned for the task, in terms of providing good reasoning compared to that of Claude 3.5-Haiku. This is intuitive as ``smarter'' models such as 4o can likely produce more convincing reasoning.

\section {Analysis \& Observations}
\label{sec:res_eval}

\begin{table}[]
\small
\centering
\caption{LLM and Human Evaluation Results}
\label{results}
\begin{tabular}{lcccc}
\toprule
\textbf{Arch.} & \textbf{Model} & \textbf{Hall} & \textbf{Correctness} & \textbf{Reasoning} \\ \midrule
BCA    & GPT-4o     & 0.023  & 0.65 & 0.21 \\
PCA1   & GPT-4o     & 0.227  & 0.75 & 0.76 \\
PCA2   & Claude-3.5 & 0.136  & 0.77 & 0.53 \\ \bottomrule
\end{tabular}
\caption*{\footnotesize Note: The hallucinated percentages were calculated based on LLM evaluations using Llama3.1-405b as the evaluator. Correctness scores (0-1) are calculated by assigning 1 for correct, 0.5 for partially correct, and 0 for not correct responses, then averaging. Reasoning scores (0-1) are calculated by assigning 1 for strongest, 0.5 for modest, and 0 for weakest reasoning, then averaging. The headings Arch. and Hall. refer to architecture and hallucinated answers, respectively.}
\end{table}

Table~\ref{results} shows our evaluation of compliance architectures (BCA and PCAs). The data show that PCA variants, in general, provide more correct responses. This increase suggests that the PCA method works better for OTCS compliance assessment in an area where there are few automated solutions.

The hallucination rates between architectures need careful reading, as different retrieval methods affect these measurements, PCA and BCA cannot be compared in this metric. Our analysis found several key differences between the tested systems.

Both PCA experiments used more accurate technical terms compared to the more general language seen with BCA. The context information often enhanced the final answers, especially when compliance criteria appeared in the provided context. This contributes to the higher correctness scores in PCA variants.

Comparison of PCA1 and PCA2 showed different behaviours. PCA2 (Claude-3.5) used very brief reasoning steps, sometimes resulting in less detailed reasoning. Yet, this brevity sometimes helped by reducing over-thinking on simple questions and leaving less room for hallucination. The choice of LLM clearly affects both the reasoning process and the final answers.

PCA1 had some drawbacks, mainly focussing too much on specific query words and tending to be stricter and more negative. This version was less willing to make connections without clear evidence, a problem that could be fixed by testing more specific queries.

A repeated issue in PCA systems was sometimes mixing up context information with document content. Though rare, these cases usually included direct mention of the context material before deciding it was not relevant to the question. This is an odd behaviour given the direct instructions in the prompt to avoid this confusion. This highlights the need for better prompt refinement to define clearer use of the different types nodes, perhaps by providing an example. This would ultimately reduce the ambiguity in the LLM's understanding of the prompt.

The key takeaway from our results and analysis is that the retrieval of the correct document chunks plays the largest part in proper reasoning and general correctness. This is particularly notable in longer queries that require a lot of information from the document retriever. This also encourages hallucination in the response. By improving the retriever's ability to filter irrelevant documents or by incorporating domain-specific embeddings, this could significantly reduce errors and improve system reliability. Since retrieval is essential to obtain accurate and relevant information, optimising this component is essential to improve overall performance.



\subsection {Limitations}

The case study used in the compliance verification analysis only partially represents what would be expected in a comprehensive suite of OTCS documentation. Specifically, it includes only an initial risk assessment, security requirements, and a definition of the initial zoning of a SuC. Furthermore, the remainder of the case study merely refers to other complementary documents which are presumed to exist. This limitation is due in part to the scarcity of detailed OTCS documentation within the rail industry and the reluctance to share sensitive cybersecurity data \cite{patwardhan2021distributed}. In addition, the document used in the presented research is significantly smaller in size than typical real-world compliance documents, where the method excels in managing large datasets. Despite these challenges, the case study serves as a valuable starting point, highlighting key OTCS issues and allowing a suitable analysis that can be inferred from the available information. As a result of the chosen case study, compliance-related questions tended to focus on the existing sections that are present in the case study.

Another limitation is the method used to analyse LLM responses. Since compliance verification is not simply a matter of checking whether something exists or not, the system must provide sufficient reasoning to extract actionable insights from OTCS documentation. This means that the involvement of a human expert in the loop is essential, although the analysis may be subjective and opinion-based.
Despite the existence of LLM-based correctness metrics, it is deemed that they are inappropriate for this task, as the evaluation requires domain-specific knowledge for a detailed performance analysis, beyond just detecting hallucinations.
It is suspected that automated systems cannot yet replicate the nuanced understanding and expertise needed to assess OTCS compliance verification systems.


Although automated evaluation methods are much faster and more cost effective, they are likely to be less accurate than human evaluation of LLM responses, particularly when it comes to identifying irrelevant information. LLM-based metrics may struggle to detect when a chunk of text, although factually correct, is not relevant to answering the question at hand. This is especially true in the OTCS domain, where nuanced understanding is crucial.

\section{Conclusions}
\label{sec:conclusion}

This study evaluated the performance of two compliance assessment architectures, the BCA and the PCA, in the Operational Technology Cybersecurity (OTCS) domain, as described in Section~\ref{sec:methodology:Baseline} and Section~\ref{sec:methodology:parallel}, by implementing a multi-stage retrieval system powered by Retrieval-Augmented Generation (RAG). The presented findings show that the retrieval system is a bottleneck in this process and plays a crucial role in ensuring accurate responses from LLMs, as described in Section~\ref{subsec:PCA_Experiment}. Although both architectures demonstrate promising results, they exhibit different strengths and weaknesses in handling compliance-related queries. The presented evaluation methodology combined LLM-based assessment (LLM-as-a-Judge) with expert human evaluations to ensure a comprehensive analysis.

The study highlights the need to optimise retrieval processes, improve prompt engineering, and explore model fine-tuning to improve the accuracy of compliance verification. In addition, integrating more comprehensive cybersecurity documentation and case studies will further validate the effectiveness of the presented approach. As LLMs continue to evolve, their potential to automate and improve cybersecurity compliance verification remains promising, paving the way for more intelligent, scalable, and reliable assessment systems.

Despite these findings, there is significant room for further experimentation and development. Given that the LLM in the presented approach is not fine-tuned and relies solely on its pre-trained knowledge, the reasoning capabilities are impressive, which approach near-human expert-like judgment using RAG. The proposed solution remains lightweight due to the lack of fine-tuning and is highly adaptable for various domains by simply augmenting the retrieved data. This flexibility enhances its practicality and scalability, making it a versatile tool for compliance verification and other complex analytical tasks beyond OTCS.

In terms of future work, optimising retrieval parameters such as the chunk size and retrieval amount based on the complexity of the query could improve performance. Furthermore, experimenting with agentic RAG to decide when sufficient document/context chunks have been retrieved to answer the query would be an interesting extension to the research.



\bibliographystyle{IEEEtran} 
\bibliography{biblio}

\end{document}